\documentclass{ametsocV6.1}

\title{Robustness Test for AI Forecasting of Hurricane Florence Using FourCastNetv2 and Random Perturbations of the Initial Condition}

\authors{
Adam Lizerbram\aff{a},
Shane Stevenson\aff{a},
Iman Khadir\aff{b},
Matthew Tu\aff{b},
Samuel S. P. Shen\aff{b}
\correspondingauthor{Samuel S. P. Shen, sshen@sdsu.edu}
}

\affiliation{\aff{a}{Department of Computer Science, San Diego State University, San Diego, California, United States}\\
\aff{b}{Department of Mathematics, San Diego State University, San Diego, California, United States}}

\abstract{Understanding the robustness of a weather forecasting model with respect to input noise or different uncertainties is important in assessing its output reliability, particularly for extreme weather events like hurricanes. In this paper, we test sensitivity and robustness of an artificial intelligence (AI) weather forecasting model: NVIDIA’s FourCastNetv2 (FCNv2). We conduct two experiments designed to assess model output under different levels of injected noise in the model’s initial condition. First, we perturb the initial condition of Hurricane Florence from the European Centre for Medium-Range Weather Forecasts (ECMWF) Reanalysis v5 (ERA5) dataset (September 13–16, 2018) with varying amounts of Gaussian noise and examine the impact on predicted trajectories and forecasted storm intensity. Second, we start FCNv2 with fully random initial conditions and observe how the model responds to nonsensical inputs. Our results indicate that FCNv2 accurately preserves hurricane features under low to moderate noise injection. Even under high levels of noise, the model maintains the general storm trajectory and structure, although positional accuracy begins to degrade. FCNv2 consistently underestimates storm intensity and persistence across all levels of injected noise. With full random initial conditions, the model generates smooth and cohesive forecasts after a few timesteps, implying the model’s tendency towards stable, smoothed outputs. Our approach is simple and portable to other data-driven AI weather forecasting models.}

\begin{document}

\maketitle

%
%
%
\statement
The robustness of an AI weather forecasting model must be quantified for forecasters to have confidence using the tool. This paper provides a robustness test of NVIDIA’s AI weather forecasting model FourCastNetv2 for the scenario of tracking Hurricane Florence in September 2018. By adding different levels of noise to the initial condition, we find that the model produces accurate hurricane paths under moderate noise, while high noise degrades accuracy but still preserves the rough trajectory. However, the model predicts a lower storm intensity whether or not influenced by the initial noise. Under completely random initial conditions, the model still generates smooth forecasts. These results suggest that FourCastNetv2 is resilient to imperfect inputs and tends to produce robust predictions.
%
%
%

%








\section{Introduction}
Inevitable noise in atmospheric measurements and nonlinear interactions among many weather variables lead to a level of uncertainty in any weather forecast model. Quantifying this uncertainty is critically important, particularly when tracking severe weather events, such as hurricanes. Would small perturbations in the initial atmospheric condition significantly change the predicted storm path and strength? In this paper, we address this kind of robustness problem by focusing on trajectory and intensity predictions of Hurricane Florence with NVIDIA’s FourCastNetv2 (FCNv2) (Bonev et al. 2023). Specifically, we evaluate the robustness of the AI weather forecasting model to imperfect initial conditions.

To illustrate the motivation of this study, Fig. 1 presents a 3-dimensional visualization of Hurricane Florence’s true wind velocity field, depicting the storm’s structure through wind speeds and directions. Figure 2 then compares the true trajectory of Hurricane Florence and track predictions generated by FCNv2 under different levels of noise applied to the initial conditions. Each panel shows the median outcome from 30 repeated runs at a given noise level, demonstrating the effect of noise on forecasts. The procedure of applying noise to the initial condition is explained in Section 2c. With little to no added noise, the model closely follows the true trajectory until the last timesteps, where it slightly deviates. With moderate noise, the start and end of the predicted track is accurate, with some displacement mid-forecast. When large amounts of noise are introduced, the prediction degrades but continues to follow the general direction of the storm. 

These two figures highlight the central question of this paper: how robust are AI weather forecasting models like FCNv2 to imperfect or unexpected initial conditions?

\begin{figure}[t]
\centering
 \noindent\includegraphics[width=22pc,angle=0]{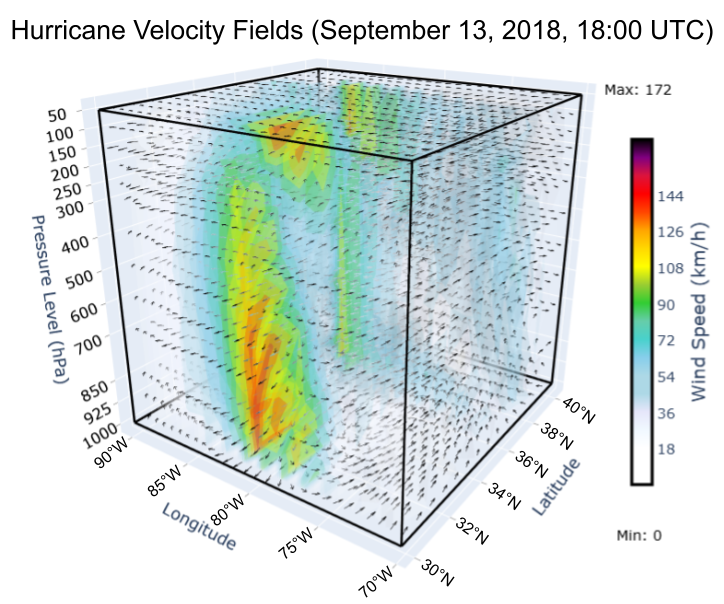}\\
 \caption{The wind velocity field of Hurricane Florence at 18:00 UTC, September 13, 2018: Wind speed (depicted by the color bar) and direction (depicted by the arrows) for the central Atlantic area (30°N–40°N latitude, 70°W–90°W longitude) over all 13 pressure levels.}\label{f1}
\end{figure}

\begin{figure}[t]
\centering
 \noindent\includegraphics[width=35pc,angle=0]{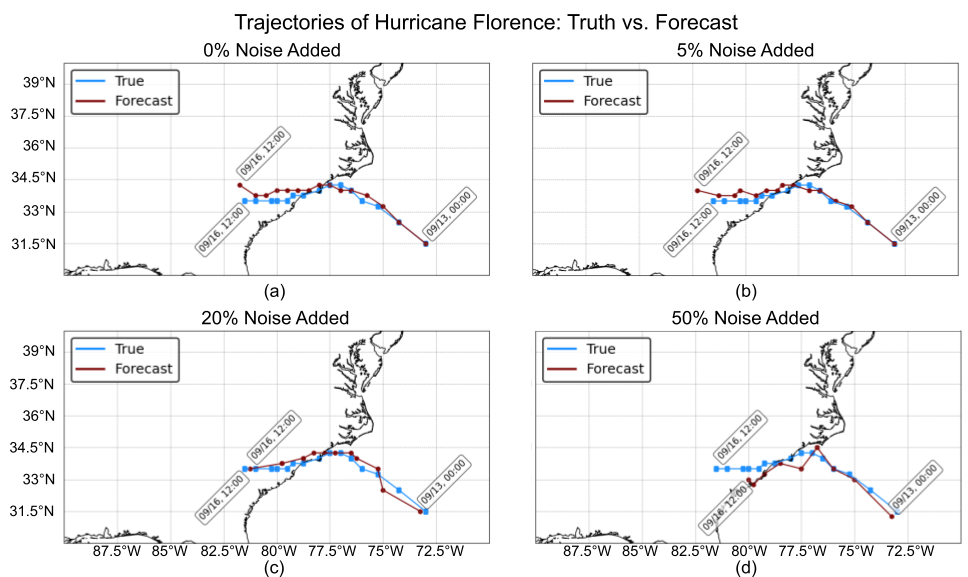}\\
 \caption{Trajectories of Hurricane Florence (September 13–17, 2018): Truth in blue, and forecast in red from FCNv2 predictions under varying levels of Gaussian noise applied to the initial condition: (a) 0\% noise added, (b) 5\% noise added, (c) 20\% noise added, and (d) 50\% noise added. Noise percentages and application procedure are described in Section 2c. Each of the shown predictions are the median outcome of 30 runs of each noise level in terms of cumulative positional error, explained in Section 2d.}\label{f2}
\end{figure}

Current operational weather forecasting models are numerical. Numerical weather prediction (NWP) models represent the mathematical equations governing atmospheric dynamics, and may be sensitive to small variations in initial conditions. Torn and Hakim (2007) aimed to quantify this uncertainty through ensemble forecasting (Wu and Levinson 2021), using the NWP model Weather Research and Forecasting (Skamarock 2005). Such ensemble-based approaches to quantifying uncertainty could be extended to evaluating modern AI weather forecasting models.

More recently, studies have also examined the performance of AI weather forecasting models in predicting extreme weather events. Charlton-Perez et al. (2024) compared several state-of-the-art AI weather models, including FourCastNet (Pathak et al. 2022), FourCastNetv2 (Bonev et al. 2023), GraphCast (Lam et al. 2023), and Pangu-Weather (Bi et al. 2022), against the NWP model Integrated Forecasting System (IFS) (Buizza et al. 2018) in tracking Storm Ciarán, an intense European windstorm in November 2023. They found that the AI models excelled in tracking the storm’s path, often matching or surpassing IFS, but consistently underestimated wind strengths. Another study by Shi et al. (2025) extended this research to the 2023 Western Pacific tropical cyclone season, evaluating IFS and Pangu-Weather. Their results similarly demonstrated that Pangu-Weather provided significant improvements in cyclone track accuracy compared to IFS, but again underestimated storm intensity. However, neither of these evaluations included the quantification of uncertainty due to imperfect initial conditions, such as noise in input data. 

In general, the data-driven AI weather models are trained exclusively on high-quality atmospheric datasets such as European Centre for Medium-Range Weather Forecasts (ECMWF) Reanalysis v5 (ERA5) (Hersbach et al. 2020). Then, are they still robust under imperfect or unexpected inputs? Roy (2024) defined the robustness of AI models as the ability to handle such inputs gracefully and produce a stable prediction. This quality is important for operational weather models since real-world data is prone to containing imperfections, such as noise or missing data. Rackow et al. (2024) explored the robustness of state-of-the-art ML weather forecasting models to out-of-distribution input, including Pangu-Weather (Bi et al. 2022), GraphCast (Lam et al. 2023), and AIFS (Lang et al. 2024), by initializing them with atmospheric states from cooler pre-industrial, modern, and warmer future climates. While the models retained skill in producing accurate forecasts under these unfamiliar conditions, they did exhibit bias toward the climate state on which they were trained.

One way to design a robustness study is to use ensemble forecasting (Wu and Levinson 2021) to generate many forecasts with slightly different perturbed initial conditions. So far, the published literatures are limited in assessing how state-of-the-art AI weather models respond to imperfect inputs in the context of hurricane trajectory prediction. This paper helps fill this gap by examining the robustness of FCNv2 to tracking hurricanes under noise-perturbed input and generating forecasts under unphysical input. In particular, we track Hurricane Florence during September 13–16, 2018 under initial conditions perturbed with varying amounts of Gaussian noise. We also test how FCNv2 handles unexpected inputs by generating forecasts from fully randomized initial states, sampled from $\chi^2$, normal, lognormal, and uniform distributions. We demonstrate that the model maintains forecast structure even under significant perturbations, and transforms random initial conditions into cohesive weather dynamics.

Section 2 describes the datasets and method used in this paper. Results are included in Section 3, and conclusions and discussion in Section 4.

\section{Data and method}

\subsection{Data}

We use the ERA5 data as the input to the FCNv2 model. We extract ERA5 fields over the timeframe of September 13 at 00:00 UTC to September 16 at 12:00 UTC, 2018, encapsulating the evolution of Hurricane Florence in the Atlantic ocean. Data is retrieved at 6 hour intervals, 0.25°-by-0.25° latitude and longitude resolution, and 13 atmospheric pressure levels, using the Copernicus Climate Data Store (CDS) Application Programming Interface (API). We use a total of 73 variables from ERA5, arranged to match the configuration of FCNv2. These variables include a mix of surface and pressure level fields, such as mean sea level (MSL) pressure, zonal and meridional wind components, relative humidity (RH), and surface pressure (SP). Table 1 lists each of the 73 variables (Khadir et al. 2025).

\begin{table}[t]
\caption{ FourCastNetv2’s 73 weather variables (Khadir et al. 2025). The first eight variables are for a single layer, and the other five variables are at 13 pressure levels: 50, 100, 150, 200, 250, 300, 400, 500, 600, 700, 850, 925, and 1000 hPa. Eight single layer variables plus 5 variables at 13 pressure levels result in a total of 73 variables.}\label{t1}
\begin{center}
\begin{tabular}{|c|c|c|}
\hline
Variable & Description & Units\\
\hline
$u_{10m}$ & Zonal wind 10 m above the surface & m/s \\
$u_{100m}$ & Zonal wind 100 m above the surface & m/s \\
$v_{10m}$ & Meridional wind 10 m above the surface & m/s \\
$v_{100m}$ & Meridional wind 100 m above the surface & m/s \\
$T_{2m}$ & Temperature 2 m above the surface & K \\
$SP$ & Surface pressure & Pa \\
$MSL$ & Mean sea level pressure & Pa \\
$TCWV$ & Total column water vapor & kg/m$^2$ \\
\hline
$z_{p}$ & Geopotential height at pressure level $p$ hPa & m \\
$T_{p}$ & Temperature at pressure level $p$ hPa & K \\
$u_{p}$ & Zonal wind at pressure level $p$ hPa & m/s \\
$v_{p}$ & Meridional wind at pressure level $p$ hPa & m/s \\
$RH_{p}$ & Relative humidity at pressure level $p$ hPa & \% \\
\hline
\end{tabular}
\end{center}
\end{table}

\subsection{Model access and prediction}

We use the pretrained FourCastNetv2 model provided through NVIDIA’s Earth-2 Model Intercomparison Project (MIP) GitHub repository (NVIDIA 2025). This open source initiative offers access to several ML weather prediction models. The Earth-2 MIP interface supplies a pretrained instance of FCNv2 and allows any custom data with compatible format as input to the model.

This study is limited to run FCNv2 predictions only and does not include retraining or fine tuning. FCNv2 follows a similar overall architecture as the original FourCastNet model (Pathak et al. 2022). The model standardizes the input data for each weather variable, and automatically rescales the running output data to match the units of the original variables. The timestep of FCNv2 is 6 hours. The model is deterministic, meaning that the same input will always result in the same output. The output of the model over all timesteps is stored for later evaluation and visualization.

\subsection{Experimental design}

To evaluate the robustness of FCNv2 to imperfections in initial conditions, we conduct two sets of experiments: (i) Injecting varying amounts of noise into the ERA5 initial condition, and (ii) fully randomizing the initial conditions. These experiments assess the model’s behavior both in operational uncertainties such as measurement error and extreme unexpected cases.

\subsubsection{Noise-injected initial conditions}

In the first set of experiments, we perturb the ERA5 initial condition from September 13, 2018, at 00:00 UTC by adding Gaussian noise independently to all grid points for each weather variable. The perturbed data is then passed through FCNv2 to generate 84-hour forecasts, or 15 timesteps at 6 hour intervals: From September 13, 00:00 UTC to September 16, 12:00 UTC, 2018. These forecasts are evaluated in two ways: (i) Tracking the eye of Hurricane Florence using the local minimum MSL pressure to analyze trajectory predictions, and (ii) pixelwise forecast error analysis to assess the model’s adaptation to varying noise levels. This noise perturbation is mathematically described as follows. 

Let $x_L$ represent the true value of atmospheric variable $x$ at latitude/longitude pair $L$, $\mu_x$ and $\sigma_x$ be the mean and standard deviation of $x$ respectively, and $\alpha,\beta \in [0,1]$ be chosen proportions. Then, the perturbed data $\Tilde{x}_L$ is calculated as the following for each variable and latitude/longitude pair:

\begin{equation}
\Tilde{x}_L = x_L + \xi_L,
\end{equation}
where
\begin{equation}
\xi_L \sim \mathscr{N}(\mu = \pm \alpha \mu_x, \sigma = \beta \sigma_x)
\end{equation}
is the Gaussian noise, $\alpha$ is a measure of mean bias, and $\beta$ is the percentage of the standard deviation. We test the following biases and noise percentages: 

\begin{equation}
\alpha,\beta \in \{0, 0.02, 0.05, 0.10, 0.20, 0.35, 0.50\}.
\end{equation}

Our tests with biased noise ($\alpha \not=0$) caused the model to produce completely divergent and nonsensical forecasts within a few timesteps; possible reasons for this will be discussed in Section 4. Thus, we restrict our analysis to zero-mean perturbations ($\alpha=0$). When the quantity of noise is referred to as a percentage, that corresponds to the value of $\beta$. For each noise level, the experiment is performed 30 times with different random seeds. The median instances of the experiments, in terms of mean trajectory error (MTE; explained in Section 2d), are visualized and studied more closely. Fig. 3 displays examples of the initial condition with varying levels of zero-mean Gaussian noise.

\begin{figure}[t]
 \noindent\includegraphics[width=35pc,angle=0]{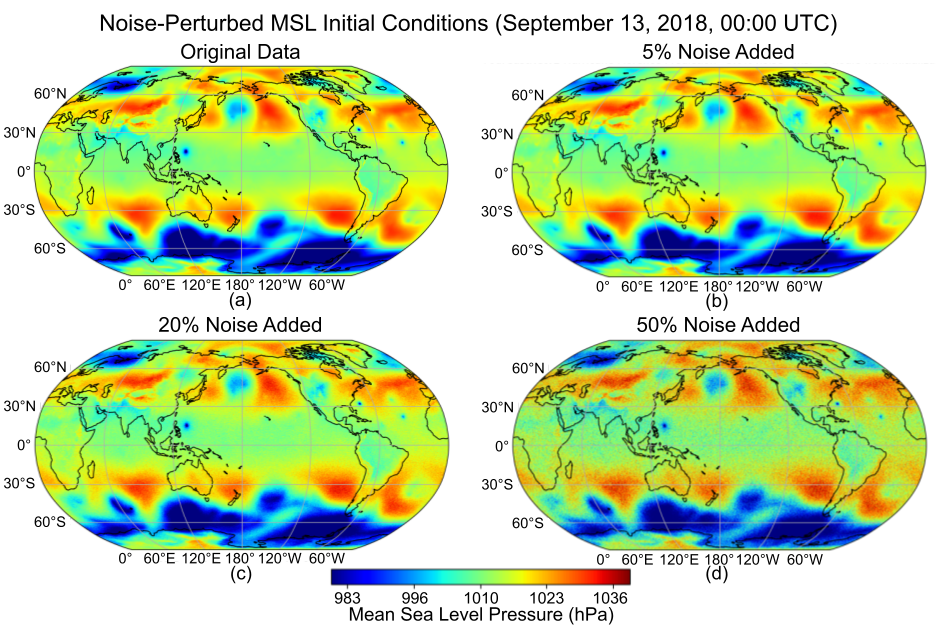}\\
 \caption{Perturbed initial conditions for mean sea level pressure (MSL) with varying levels of noise added: (a) Original ERA5 data (0\% noise added), (b) 5\% noise added, (c) 20\% noise added, and (d) 50\% noise added.}\label{f3}
\end{figure} 

\subsubsection{Fully randomized initial conditions}

In the second set of experiments, we generate random initial conditions by sampling from several distribution types (normal, lognormal, uniform, $\chi^2$) across all weather variables and grid points, scaled to match the distributions of the ERA5 data. These random states are not physically realistic but test how the model behaves given such unexpected data. No constraints on real atmospheric states are applied. Figure 4 portrays the fully random initial conditions sampled from each of the tested distributions.

\begin{figure}[t]
 \noindent\includegraphics[width=30pc,angle=0]{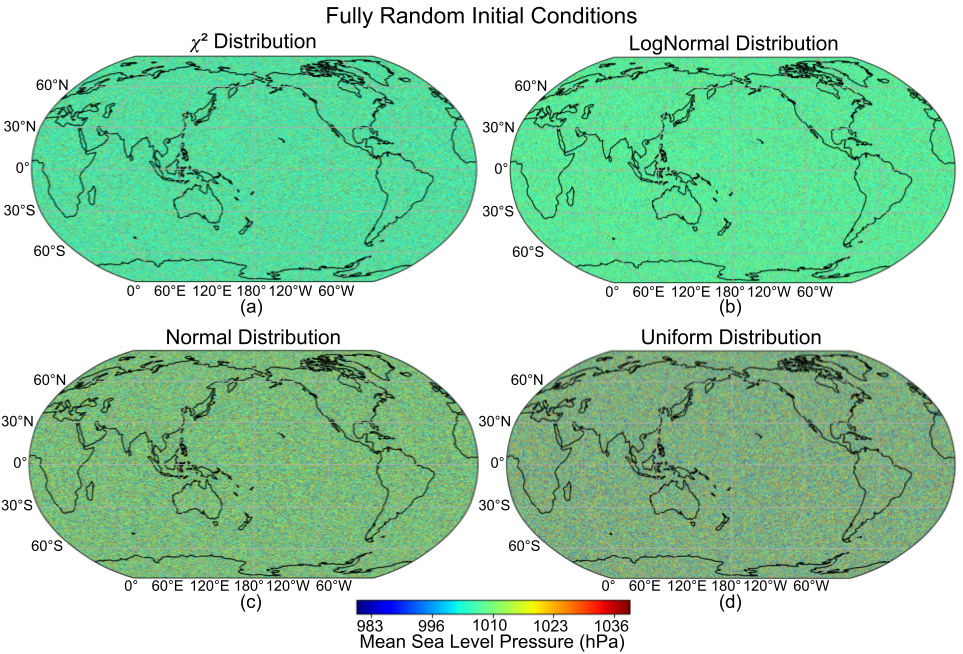}\\
 \centering
 \caption{Fully random initial conditions generated from different distributions: (a) $\chi^2$, (b) lognormal, (c) normal, and (d) uniform. The fields appear primarily green because most sampled values are near the center of the color bar. Although this particular color bar is for MSL, these fully random initial conditions can be generated for any of the 73 variables.}\label{f4}
\end{figure}

These experiments are used for global analysis only. Since the initial conditions do not contain a storm system, attempting to track a storm would not provide meaningful results. Forecasts are again generated for 15 timesteps. We assess whether FCNv2 is able to “forget” the unrealistic initial conditions and produce spatially coherent, plausible results after several timesteps.

Together, these two experiments allow us to analyze FCNv2’s ability to handle both realistic perturbations and extreme unphysical inputs.

\subsection{Evaluation metrics}

For regional analysis (noise-perturbed initial condition experiment only), we evaluate the accuracy of FCNv2’s Hurricane Florence trajectory predictions by computing the mean trajectory error (MTE), measured in kilometers, between the predicted and true hurricane trajectories over the 84-hour forecast period. The hurricane center is identified using the minimum MSL pressure at each timestep. The cumulative sum of the distances between the eye of the storm in the real and predicted data, normalized by the number of timesteps, provides a measure of tracking error under each noise level. For all weather variables $x$ and latitude/longitude pairs $L$ restricted to the central Atlantic area (30°N–40°N latitude, 70°W–90°W longitude), we also calculate pixelwise mean biased error $\varepsilon_x(L)$ between the forecast $f_x(L)$ and true $t_x(L)$ data as
\begin{equation}
\varepsilon_x(L) = f_x(L) - t_x(L),
\end{equation}
meaning that positive error is an overestimate in the prediction and vice versa. We specifically focus on MSL pressure errors (choose $x=MSL$). With this data, we (i) generate histograms of the initial noise and final-timestep error distributions, (ii) plot the evolution of quantile statistics (max, 95th, 75th, median, 25th, 5th, min), and (iii) track changes in statistical moments (mean, standard deviation, skewness, kurtosis) over timesteps and noise levels. These metrics help us understand the shape and magnitude of the error distributions.

For global analysis (both noise-perturbed and fully random initial condition experiments), we evaluate model performance using the same mean biased error analyses without restricting the latitude/longitude pairs $L$ to the central Atlantic. In this case, we focus on histograms to track how the global error distributions evolve over time.

This combination of trajectory- and distribution-based error metrics allows us to evaluate both the hurricane tracking-specific and overall global robustness of FCNv2 to imperfect initial conditions.

\section{Results}

The primary focus of our analysis is to evaluate FCNv2’s ability to preserve the structure and trajectory of Hurricane Florence under perturbations in the initial condition. To complement these results, we also conduct experiments on a global scale to test how forecast errors evolve under the same perturbations, and how the model behaves under fully random, unphysical inputs.  

\subsection{Regional analysis: Hurricane Florence tracking under noisy initial conditions}

To evaluate the robustness of FCNv2, we examined its ability to track Hurricane Florence under varying levels of Gaussian noise applied to the initial condition. The storm center was tracked using the position of the minimum MSL pressure. Figure 2, introduced earlier, provides a qualitative preview of the model’s ability to track Florence under noisy initial conditions. The supplemental animations offer a dynamic visualization of forecasted MSL pressure (depicted by the background heatmaps) and wind vectors (depicted by the arrows, using $u_{10m}$ and $v_{10m}$ values) in comparison to the true data (see the animation URL in the data availability statement). Here, we expand on these qualitative observations with quantitative analyses of MTE and pixelwise mean biased error distributions of MSL pressure. MTE was calculated as the cumulative sum of distances between the predicted and true hurricane position and normalized by the number of timesteps, measured in kilometers. Pixelwise errors were calculated as true subtracted from predicted MSL pressure for each latitude/longitude pair.

Figure 5 presents the relationship between noise level and distribution of mean trajectory error over 30 trials. For up to 10\% noise added, the predicted hurricane trajectory tends to accurately follow the true storm path, with low and stable MTE. As noise levels rise to 20\% or above, the predicted trajectories begin to increasingly deviate from the true path. Despite this, even at the highest tested noise level of 50\%, the predicted hurricane track generally follows the correct direction.

\begin{figure}[t]
 \noindent\includegraphics[width=20pc,angle=0]{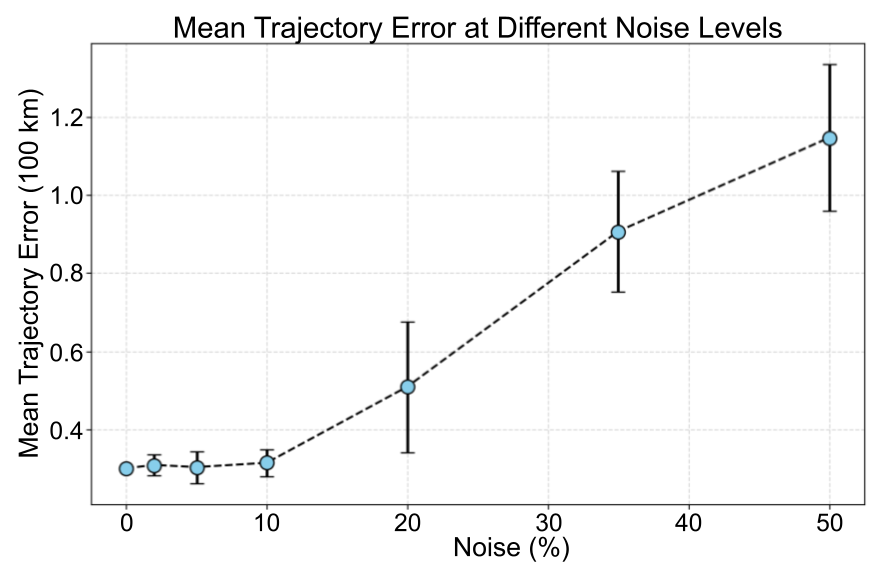}\\
 \centering
 \caption{Mean trajectory error ($\times$ 100 km) as a function of noise level (\%) with the vertical bars indicating the standard deviation from 30 trials.}\label{f5}
\end{figure}

In the supplemental animations, by comparing the true and predicted MSL heatmaps, we observe a consistent underestimation of storm longevity and intensity in each forecast under all perturbation levels, regardless of the accuracy of the predicted hurricane track. To examine this behavior more closely, we analyzed the evolution of the mean biased error distributions of MSL pressure throughout the forecast (Figs. 6–8).

The histograms in Fig. 6 provide a qualitative view, with the error distribution widening over time for lower noise levels (0–5\% noise added), and narrowing over time for moderate to high noise levels (20–50\% noise added). In all cases, the error distributions are consistently right-skewed by the end of the forecast. 

\begin{figure}[t]
\centering
 \noindent\includegraphics[width=28pc,angle=0]{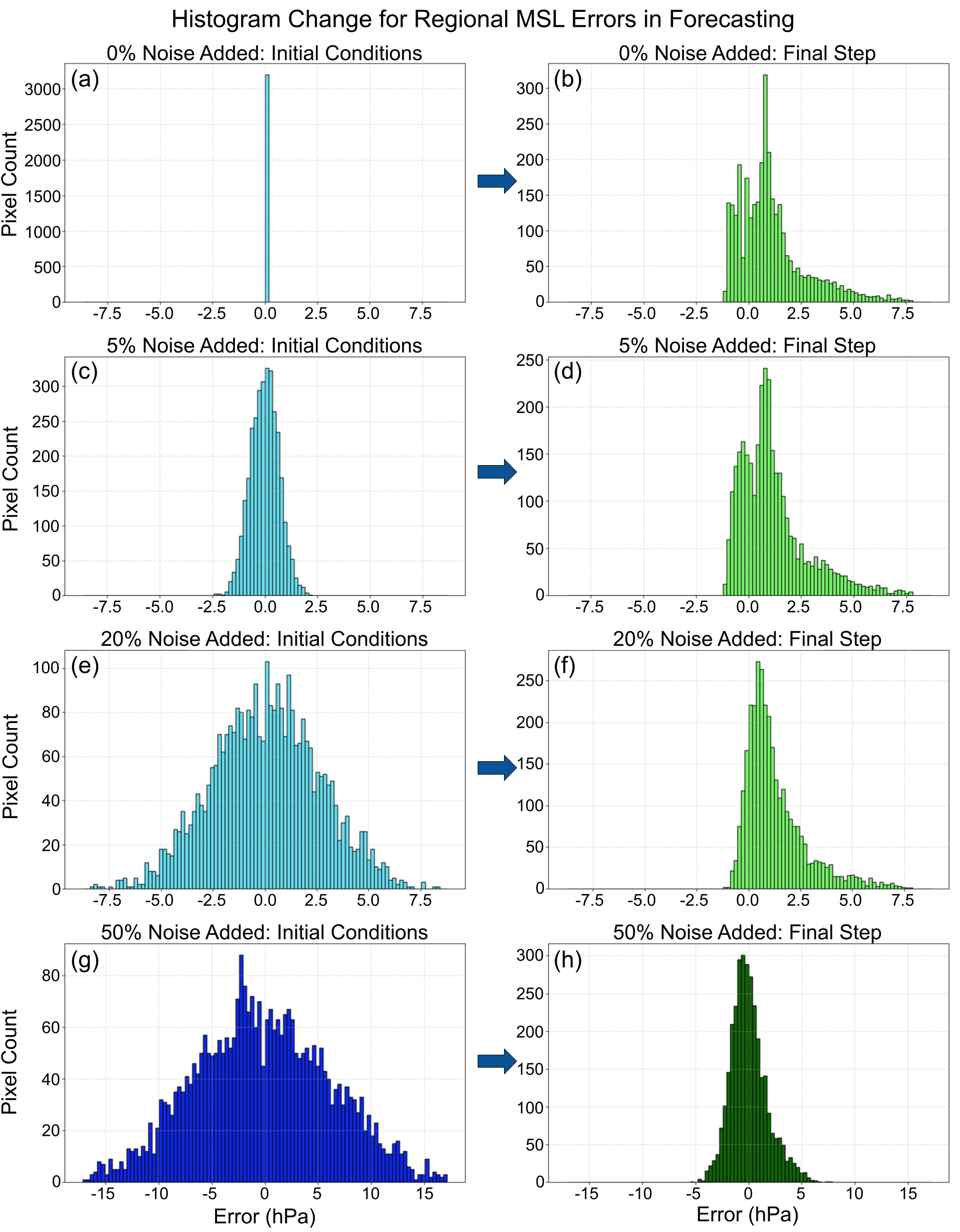}\\
 \caption{Left panels show the histograms of the noise added to the initial conditions. Right panels show the histograms of the mean biased errors at the final forecast timestep (84 hours after initialization). Panels (a) and (b) are for 0\% noise added (original data), (c) and (d) for 5\% noise added, (e) and (f) for 20\% noise added, and (g) and (h) for 50\% noise added. Panels (g) and (h) are darker colors to emphasize the larger error scale between ±15 hPa instead of the ±7.5 hPa scale for panels (a)–(f).}\label{f6}
\end{figure}

The change in quantile statistics over time in Fig. 7 quantitatively represents how the error distribution evolves over time. For low noise levels (0–5\%), the error distributions start narrow, widen mid-forecast, then narrow slightly by the end but remain broader than at the start. For higher noise levels (20–50\%), they start wide, broaden further, and contract to a final shape narrower than the initial. The median error is consistently positive for 0–20\% noise added, but fluctuates near zero depending on the particular initial perturbation for 50\%. Across all cases, however, the maximum positive errors are consistently larger in magnitude than the negative errors.  

\begin{figure}[t]
 \noindent\includegraphics[width=35pc,angle=0]{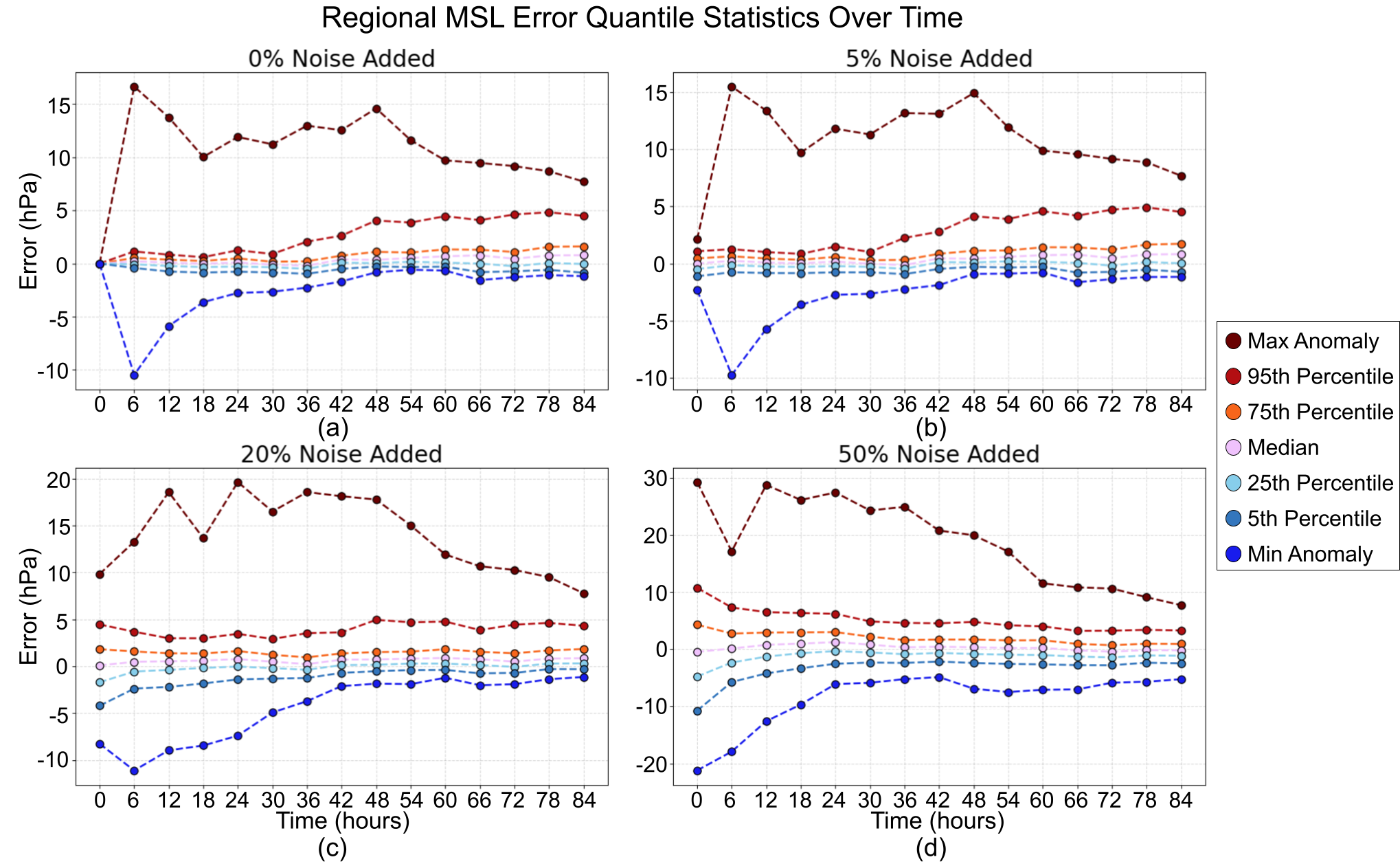}\\
\centering
 \caption{Time series of quantile statistics for the errors of MSL pressure at different noise levels: (a) 0\% noise added, (b) 5\% noise added, (c) 20\% noise added, and (d) 50\% noise added.}\label{f7}
\end{figure}

The moment analysis in Fig. 8 shows that the average error steadily increases across the forecast for 0–20\% noise added. For 50\% noise added, error increases in the early stages, peaks mid-forecast, and decreases to near-zero in the later stages. The standard deviation increases over time for low noise cases but decreases over time for higher noise, consistent with the widening and narrowing behaviors observed in Fig. 7. Skewness and kurtosis both remain positive in all cases, increasing until mid-forecast, then decreasing in the end stages, with skewness remaining positive and kurtosis approaching zero.

\begin{figure}[t]
\centering
 \noindent\includegraphics[width=30pc,angle=0]{fig08.png}\\
 \caption{Time series of four statistical indices (mean, standard deviation, skewness, kurtosis) of MSL pressure errors at different noise levels: (a) 0\% noise added, (b) 5\% noise added, (c) 20\% noise added, and (d) 50\% noise added.}\label{f8}
\end{figure}

Together, Figs. 5–8 show that MTE remains stable at low noise but grows sharply beyond 20\% noise added, while pixelwise MSL pressure errors are consistently right-skewed. The following section expands this analysis to global scales.

\subsection{Global analysis: forecast stability under noisy and fully random initial conditions}

To assess the global sensitivity of FCNv2 to perturbed inputs, we (i) injected Gaussian noise of a variety of magnitudes into real initial conditions from the ERA5 dataset, and (ii) generated forecasts from fully random initial conditions sampled from various distributions. For both experiments, we analyzed mean biased errors for MSL pressure over time, calculated the same way as in the previous section. We focus first on the noise injection experiments.

We computed histograms of pixelwise MSL errors for the first and last timesteps of the forecast, depicted in Fig. A1. For low noise levels (0–5\% noise added), the error distributions gradually spread over time, while under moderate to high noise perturbations (20–50\% noise added), the error distributions narrowed over time. Despite the different perturbation magnitudes, the global-scale errors converged toward similar zero-centered distributions. For moderate perturbations ($\le$ 20\% noise added), 99\% of errors fell within ±7.5 hPa, while for the highest perturbation level (50\% noise added), 99\% were within ±15 hPa.

We next evaluate forecasts generated from fully random initial conditions. In this experiment, random initial conditions were sampled from several different distributions ($\chi^2$, lognormal, normal, and uniform), and forecasts were analyzed using the same mean biased error metrics. To assess how the model responds to such unrealistic initial states, we first inspect the evolution of forecasts from random initial conditions over time. Figure 9 shows that, across each tested distribution the initial conditions were sampled from, the MSL pressure forecasts evolve into spatially coherent fields within 84 hours. 

\begin{figure}[t]
 \centering
\noindent\includegraphics[width=35pc,angle=0]{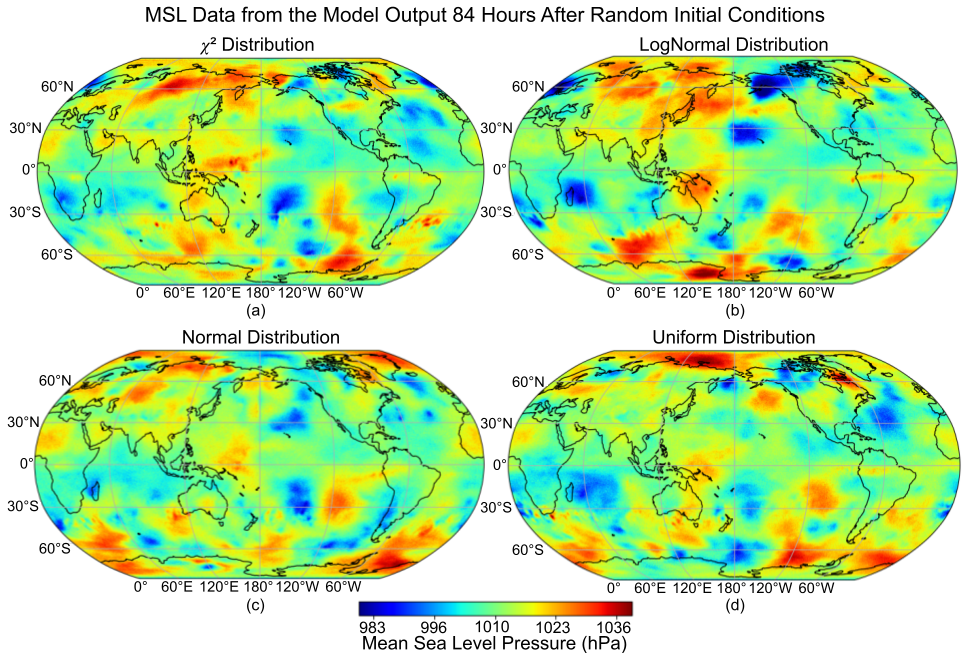}\\
 \caption{FCNv2-forecasted MSL pressure after 84 hours from fully random initial conditions with different noise distributions: for (a)  $\chi^2$, (b) lognormal, (c) normal, and (d) uniform. For the corresponding initial conditions, see Fig. 4.}\label{f9}
\end{figure}

However, this behavior is not consistent across all forecasted variables. Our analysis has focused on MSL pressure since it is used for tracking Hurricane Florence in the regional experiments, but FCNv2 produces forecasts for 72 other variables. Most fields evolve into smooth, coherent forecasts from random initial conditions. However, surface pressure (SP) was the only field we examined that retained much of the initial randomness throughout the forecast period. Furthermore, relative humidity (RH) did produce spatially coherent patterns, but did not adhere to the physical 0–100\% range, since several negative values present in the initial condition were retained. Figure 10 illustrates the results of an 84-hour forecast from normally distributed random initial conditions for these outlier weather variables.  

\begin{figure}[t]
\centering
\noindent\includegraphics[width=35pc,angle=0]{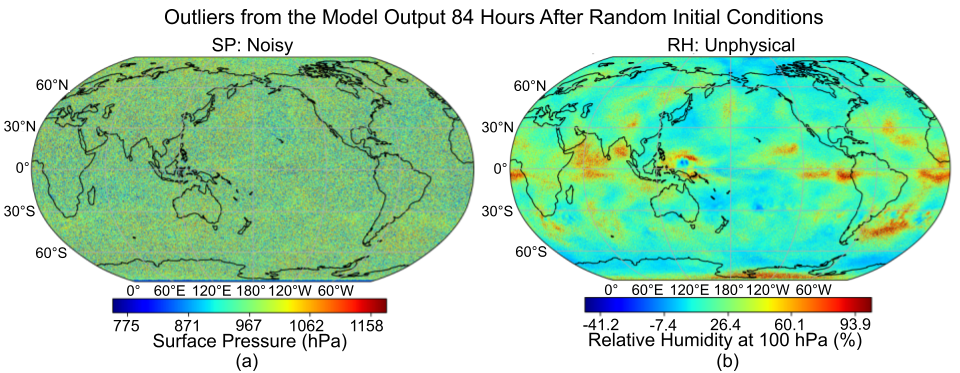}\\
 \caption{Outliers and unphysical FCNv2 outputs generated from fully randomized initial conditions after 84 hours. (a) Surface pressure (SP) retains most of the initial randomness although some slight patterns begin to emerge. (b) Relative humidity (RH) at 100 hPa develops coherent spatial patterns but includes values outside the physical 0–100\% range (negative values in blue).}\label{f10}
\end{figure}

In addition to the spatial fields, we also examined the evolution of the mean biased error distributions of MSL pressure, shown in Fig. A2. The initial distributions are broad and strongly skewed. Over time, however, these error distributions converge toward 0 both in terms of average and having a narrower spread, with all four random initializations converging to a similar near-zero-centered shape by the final timestep.

\section{Conclusions and discussion}

In this section, we first consider overall robustness in terms of hurricane tracking and global forecasting under imperfect initial conditions. Then, we discuss FCNv2’s tendency toward smooth forecasts and how this affects representation of extreme weather events. Finally, we suggest implications of model robustness for ensemble forecasting and operational use, and conclude with directions for future work.

Increasing Gaussian noise in the initial conditions progressively distorts finer-scale weather structures while mostly preserving larger-scale patterns (Fig. 3). Hurricane Florence, as a smaller-scale structure, is significantly affected by large perturbations. Small perturbations ($\le$ 10\% noise added) generally do not destabilize the hurricane structure and track, as the initial structure is mostly intact (Fig. 5). Larger perturbations ($\ge$ 20\% noise added) significantly increase MTE and reduce tracking accuracy, though the predicted tracks still tend to follow the correct overall direction. This could indicate that the model implicitly preserves hurricane motion even when the structure is heavily distorted. The pixelwise error distributions of MSL pressure also reveal a contrast between low and high noise level behaviors. With low noise, the error distributions broaden over time as expected from iterative predictions. At higher noise levels, they instead narrow over time (Fig. 6 and Fig. A1), implying a tendency to predict more common MSL pressure values. MSL pressure is consistently overestimated near the eye of Florence, reinforcing a tendency to underestimate storm intensity, aligning with what Charlton-Perez (2024) and Shi et al. (2025) presented. High kurtosis mid-forecast (Fig. 8) indicates disproportionately large errors near the center of the cyclone compared to its surroundings, which can be observed in the supplemental animations.

When initialized from fully random states, FCNv2 gradually “forgets” the incoherent inputs, imposing its learned weather patterns to produce smooth and coherent forecasts within 84 hours. This is also apparent in the error histograms: All four initial distributions, starting broad due to the random initial conditions having no resemblance to the real ERA5 data, converge to a similar near-zero-centered shape by the final timestep (Fig. A2). Most of the predicted weather variables follow this pattern, evolving over time into smooth and physically reasonable fields. However, not all variables behave consistently. For example, surface pressure (SP) retains much of the input randomness, with only faint patterns becoming visible at the end of the forecast. Relative humidity (RH) develops coherent patterns but does not correct the unphysical values outside the 0–100\% range which were present in the initial conditions. These variable-specific behaviors show that while the model strongly regularizes most fields to appear physically plausible, it does not always enforce physical consistency, allowing some unrealistic aspects from the initial conditions to persist.

Taken together, these experiments show that FCNv2 is robust in several aspects. It maintains accurate hurricane tracks under moderate perturbations, actively reduces forecast error under significant noise, and produces mostly coherent forecasts from fully randomized inputs. However, the robustness of FCNv2 is limited. Biased perturbations, unlike zero-mean noise, often cause forecasts to diverge entirely. A possible explanation for this is the model’s standardization of the initial conditions. If the mean or standard deviation of the input data does not match the statistics provided through the Earth-2 MIP interface, the standardized data will not be zero-centered. During a prediction, this offset could propagate through the neural network into exploding outputs. Other robustness limitations are variable-specific. Most weather variables eventually “forget” randomness in the initial conditions, but certain outliers such as SP and RH can never forget the unphysical nature of the input. More broadly, FCNv2 appears to approximate physical behavior under perfect initial conditions, but may favor imposing its learned weather patterns instead of adhering to physical laws under imperfect inputs. These behaviors align with the findings of Dueben and Bauer (2018), who showed that purely data-driven ML weather forecasting models can implicitly learn the dynamics of a system without explicitly encoding the physical laws into the design of the model. However, as our results suggest, the absence of physical constraints may lead to unphysical behavior under imperfect input. Integrating physical laws into data-driven AI weather forecasting models, as explored by Kashinath et al. (2021), may help maintain physicality and further improve robustness to imperfect initial conditions.

These results also have implications for ensemble forecasting and operational reliability. FCNv2’s hurricane track predictions under small perturbations, on average, perform as well as the unperturbed case in terms of MTE, while capturing plausible variations in hurricane motion (Fig. 5). This connects naturally to ensemble forecasting, where a collection of slightly perturbed forecasts may produce more reliable outcomes than a single deterministic forecast. The strong regularizing tendency observed in both noisy and random initialization experiments means that FCNv2 tends to dampen variability and converge toward stable states. Although this tendency contributes to robustness to imperfect initial conditions, it may also suppress sensitivity to finer features. For operational forecasting, these tendencies imply that FCNv2 may deliver consistent global weather patterns under imperfect initial conditions but may misrepresent intensity and longevity of extreme weather events.  

Future work could extend these experiments beyond FCNv2 and Hurricane Florence to assess whether the same behaviors appear across other AI weather forecasting models and different storm events. Another direction is to analyze forecasts produced from partial or interpolated data, representing the often incomplete information in real-time operations. The robustness of physics-informed AI models to imperfect initial conditions could be tested and compared to that of purely data-driven AI models. Finally, testing simple safeguards such as clamping variables with physical bounds (for instance, restricting RH to the 0–100\% range) could help ensure physical consistency while retaining the benefits of learned dynamics.

Overall, these experiments show how robustness in AI weather models can be evaluated by testing performance under imperfect initial conditions. By this standard, FCNv2 proves robust, consistently producing accurate hurricane tracks under noisy input and generating coherent forecasts from incoherent initial conditions, although it still struggles in representing intensity and longevity of storms and enforcing physical realism.

%

%

\clearpage
\acknowledgments
This work was partially supported by a U.S. National Science Foundation  grant (Award No. IIS 2324008) and  U.S. NOAA grants (Award No. NA22SEC4810016 and NA24NESX405C0006-T1-01).

%
%
\datastatement
The primary code and data used for visualizations in this study is available at \url{https://github.com/Alazerbeam/FourCastNetv2-Robustness-Test}. The original ERA5 data can be accessed through the Copernicus Climate Data Store (Hersbach, 2023a and Hersbach, 2023b). The version of Earth-2 Model Intercomparison Project used can be found at \url{https://github.com/Alazerbeam/earth2mip-fork}. 

%

\appendix

Figure A1 provides the global histograms of the noise in the initial conditions and the MSL pressure mean biased error at the final timestep. There are patterns parallel to the regional analysis of the error distributions widening with low initial noise and narrowing with high initial noise. Figure A2 presents the global histograms of the fully random initial distributions and the error at the final timestep. 

\begin{figure}[t]
\centering
 \noindent\includegraphics[width=25pc,angle=0]{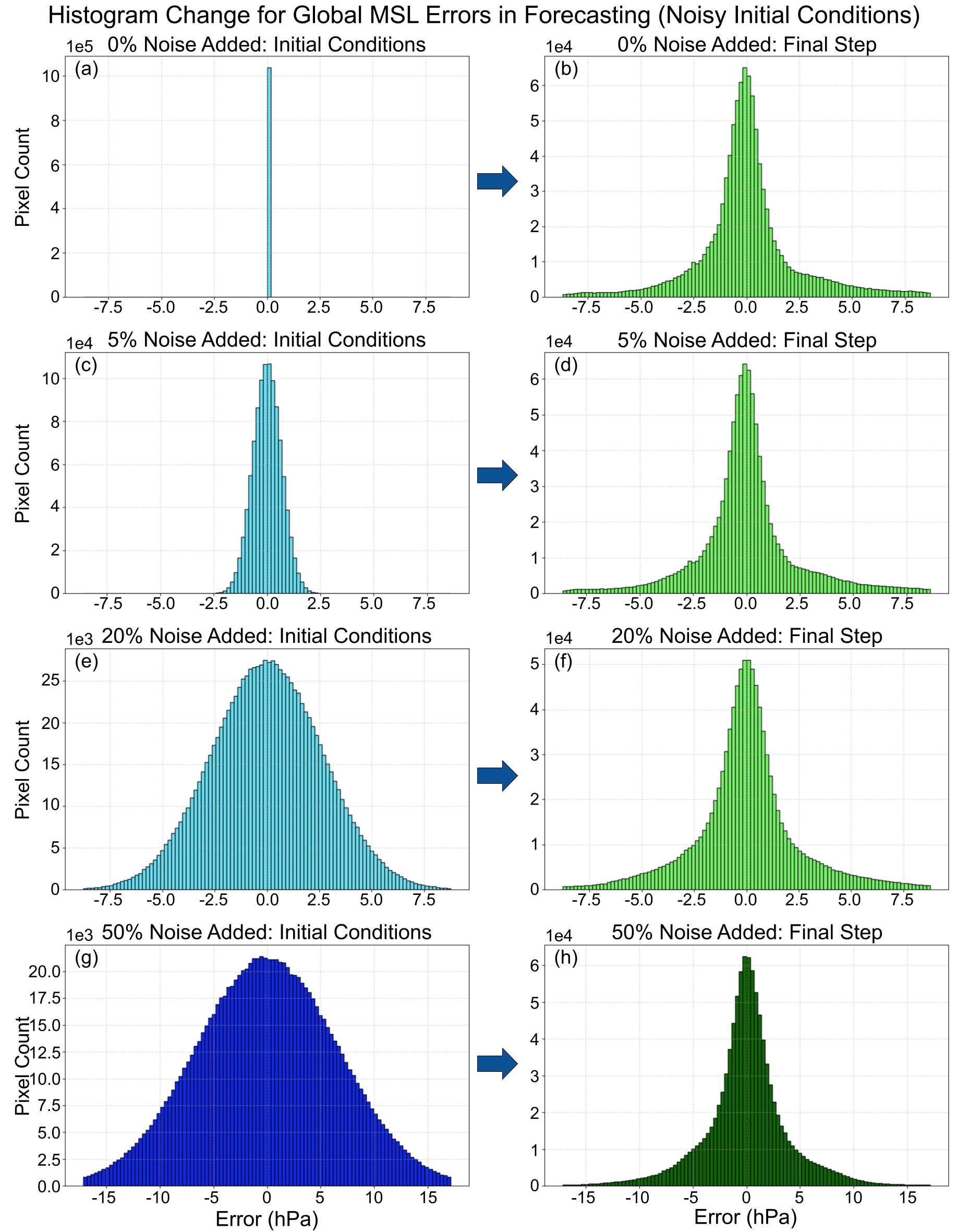}\\
 \caption{These histograms show global pixelwise mean sea level pressure (MSL) forecast error histograms for noise-injected initial condition experiments. Left panels show the histograms of the noise added to the initial conditions, while right panels show the histograms of the MSL mean biased errors at the final forecast timestep (84 hours after initialization). Panels (a) and (b) are for 0\% noise added (original data), (c) and (d) for 5\% noise added, (e) and (f) for 20\% noise added, and (g) and (h) for 50\% noise added. Panels (g) and (h) are again darker colors to emphasize the different error scale.}\label{fa1}
\end{figure}
\clearpage
\begin{figure}[t]
\centering
 \noindent\includegraphics[width=25pc,angle=0]{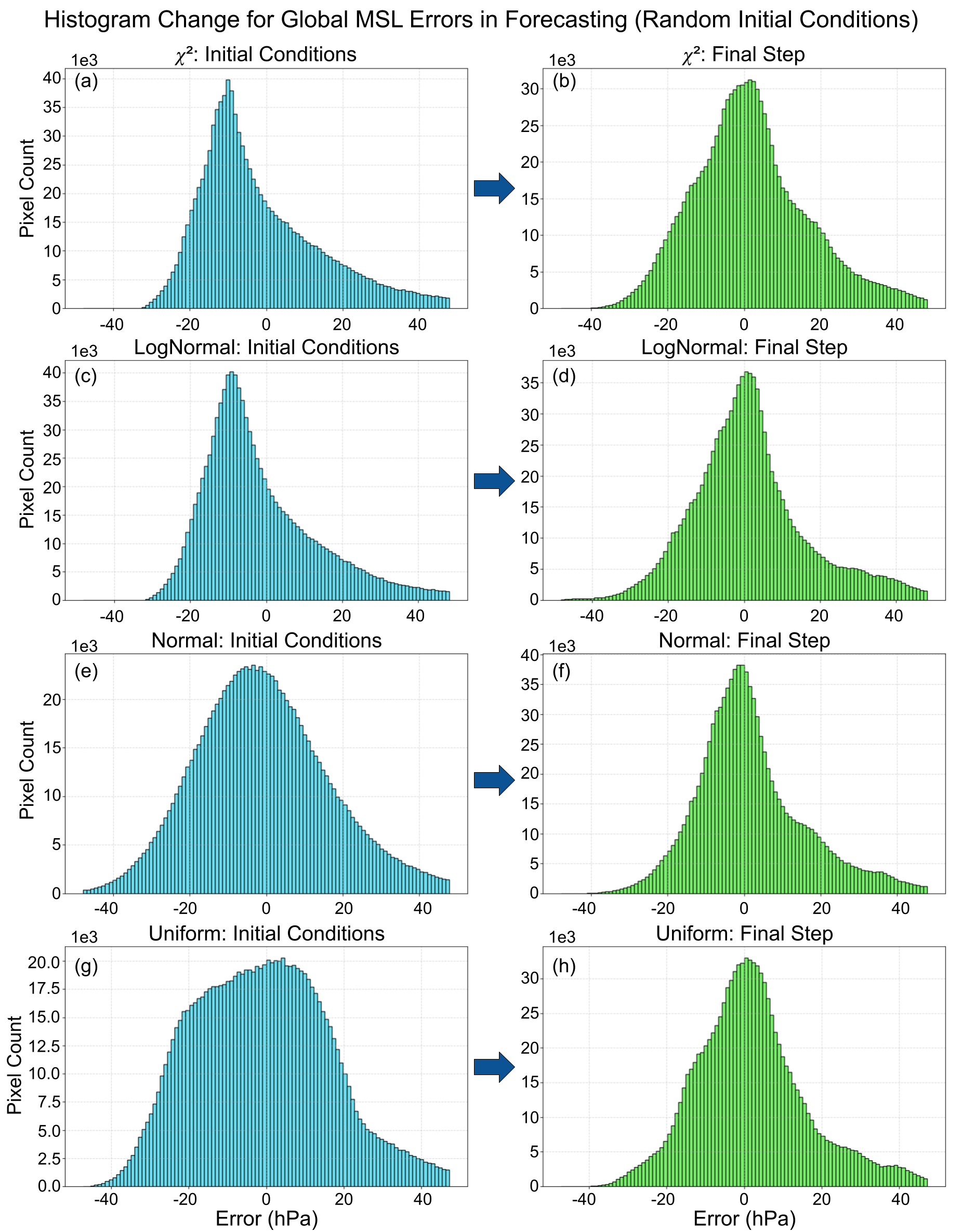}\\
 \caption{These histograms show global pixelwise mean sea level pressure (MSL) forecast error histograms for fully random initial condition experiments. Left panels show the histograms of the noise added to the initial conditions, while right panels show the histograms of the MSL mean biased errors at the final forecast timestep (84 hours after initialization). Panels (a) and (b) are for initial conditions generated from the $chi^2$ distribution, (c) and (d) for lognormal distribution, (e) and (f) for normal distribution, and (g) and (h) for uniform distribution.}\label{fa2}
\end{figure}
\clearpage





%



\newpage
\bibliographystyle{ametsocV6}
\bibliography{references}
\nocite{*}

\end{document}